# A Hybrid Anomaly Detection Method for Multivariate Time Series

**Ali Babalhavaeji[1], Hui Song[1], Mahdi Jalili[1]**

[1]RMIT University, Melbourne, Australia
Corresponding Author: Ali Babalhavaeji

## Abstract

**Multivariate time series are ubiquitous in monitoring smart systems such as power plants, smart houses, and servers. These systems use multiple sensors to collect the system's status over time. Those sensors represent the status of the whole system as one entity. Anomaly detection in such smart systems is of critical importance to prevent system failure and maintain functionality. But anomaly detection in such multivariate time series data is very challenging since this type of data includes both temporal and spatial information. Capturing both of these pieces of information is important to detect anomalies. In this paper, we propose a hybrid deep learning-based anomaly detection method, which captures both temporal and spatial information. We propose to take advantage of the feature extraction capabilities of 2d convolutional network in a variational autoencoder, alongside a prediction-based deep neural network of 2d convolutional networks and a gated recurrent unit network to capture both spatial and temporal information. Experiments on public data against state-of-the-art show the potential of our method.**

## 1- Introduction

Complex systems like power plants, modern factories, smart houses, spacecrafts, server machines, robot-assisted systems, engines, etc are monitored via smart meters (sensors). These smart meters generate vast amounts of multivariate time series data, which are complex and difficult to analyse [1]. The operators of such systems need to know whether an abnormal state occurred or not. These abnormal states indicate a fault that can be very harmful, especially for critical systems like power plants and spacecrafts [2]. Previously, such monitoring used to be done by experts who supervise the system. However, the complexity of modern IT systems and the large number of sensors to monitor the system have rendered expert-based supervision useless and call for automated approaches to detect such anomalies [3]. This problem of detecting abnormal states in historical data collected by multiple sensors is called anomaly detection in multivariate time series (MTS). The problem is an active research line [4].

This problem has complex challenges. First, MTS datasets are usually highly imbalanced, i.e., the percentage of anomalies is so low that it makes it difficult to train a model that is not biased toward normal points. Second, in the real application, usually there is no label for the data, which raises the need for unsupervised methods to deal with the problem [5]. The third challenge is the dimensions of the MTS data. In such data, the features correlate with each other and correlate over time. It is very important to consider these relationships to capture the true system status. This relation between each feature and its lagged values (auto-correlation/temporal correlation), and the inter-relation of features over time (spatiotemporal correlation), is an important piece of information and should be captured. Most of the works in the literature fail to capture all types of correlation for MTS anomaly detection explicitly. The other challenge is the stochastic nature of time series data, which further complicates anomaly detection for MTS data.
The high complexity of MTS data and the weakness of traditional methods due to the curse of dimensionality necessitate the need for deep learning-based methods for the problem. Deep-learning-based methods are categorised as prediction-based and reconstruction-based. Hundman et al [2] proposed a prediction-based method that uses Long-Short Term Memory (LSTM). LSTM proved successful in predicting time series data. Their method uses prediction error to detect anomalies. They apply their method to MTS collected by satellite devices. Tariq et al [6] propose to use a convolutional LSTM integrated with probabilistic Principle Component Analysis (PCA) as a prediction-based method.
Ikeda et al [7] developed a variational auto-encoder (VAE) for the problem. They use a multi-layer perceptron (MLP) as the building blocks of the VAE. The encoder uses a multivariate Gaussian with a diagonal covariance. The VAE is trained on normal data, and then they use reconstruction error to detect anomalies. Another line of work based on reconstruction is inspired by Generative Adversarial Networks (GAN). GAN is designed to train generative AI models by adversarial training. It proved successful in image processing to generate realistic images [8]. Li et al [1] propose a GAN-based anomaly detection for MTS data called MAD-GAN. They use LSTM for both the generator and discriminator components of the GAN. The generator builds fake MTS and passes them to the discriminator to detect fake (generated MTS) from real MTS. Although effective, GAN suffers from mode collapse and non-convergence issues [9]. Audibert et al [3] propose Unsupervised Anomaly Detection (USAD) to address these issues in GAN-based methods. They developed an Auto-Encoder (AE) architecture that consists of one encoder component and two decoder components ($AE_1$ and $AE_2$). In the first stage, the encoder and two decoders are trained to reconstruct the normal data. In the second stage, in an adversarial learning manner, $AE_2$ learns to distinguish between real data and reconstructed data by $AE_1$, and $AE_1$ is trained to fool $AE_2$ at the discrimination task. They use MLP for building blocks, and the latent space of their AE is not stochastic.
Zhang et al [10] propose a Multi-Scale Convolutional Recurrent Encoder-Decoder (MSCRED). It first constructs signature matrices for each time step. Then, a convolutional encoder captures spatial correlation, and an attention-based convolutional LSTM captures temporal correlations. Finally, a convolutional decoder is used to reconstruct the MTS. Su et al [11] propose OmniAnomaly, a VAE-based method that uses a Gated Recurrent Unit (GRU) network for building blocks of the VAE. Their method aims to capture both the temporal and stochastic nature of MTS. They use stochastic variables alongside the planar Normalizing Flows technique to learn the non-Gaussian posterior distribution of the latent space in VAE.
Previous works fail to consider all types of correlation in MTS. In this paper, we propose HEADMTS (Hybrid Ensemble Anomaly Detection Multivariate Time Series), which is a hybrid of prediction-based and reconstruction-based methods. It includes two main components. The main building blocks of these components are 2D-convolutional networks that we believe have the potential to capture correlation between different time series. The first component captures temporal correlation, and the second captures spatio-temporal correlations explicitly. We use a hybrid loss function to train the model. Our contributions are as follows:

- We propose to capture correlations between sensors explicitly using separate components.
- We propose a hybrid anomaly detection method which uses 2d-CNN as the building blocks
- We provide a comparison in terms of all important metrics.

The rest of this paper is organized as follows. In Section 2, we introduce the problem definition and the method. Section 3 includes data description, data analysis, results, and discussion. The results and discussion section includes the performance, ablation study, and the convergence rate of the method. This paper is concluded in Section 4.

# 2- Methodology

## 2-1- Problem Definition

Let's denote the input with $x \in S^{N\times M}$ , where N is the number of samples over time, and M is the number of sensors to read the MTS. Figure 1 shows the architecture of MTS. Let's denote the output for $x_t$ with $\Upsilon_t \in \{0,1\}$, where 0 indicates a normal point, and 1 indicates an anomaly. Therefore, the problem of MTSAD becomes learning $\phi: S^{W\times M} \rightarrow \{1,0\}$ where $W$ is the length of the time window of the input smart meter data. Based on the definition of the anomaly detection problem, we express the MTSAD problem as follows:

$$\phi(x_t) \rightarrow \begin{cases} 1 & if \phi(x) > \delta \\ 0 & o.w \end{cases} \tag{1}$$

where $\delta$ is a threshold that distinguishes between normal and abnormal points.

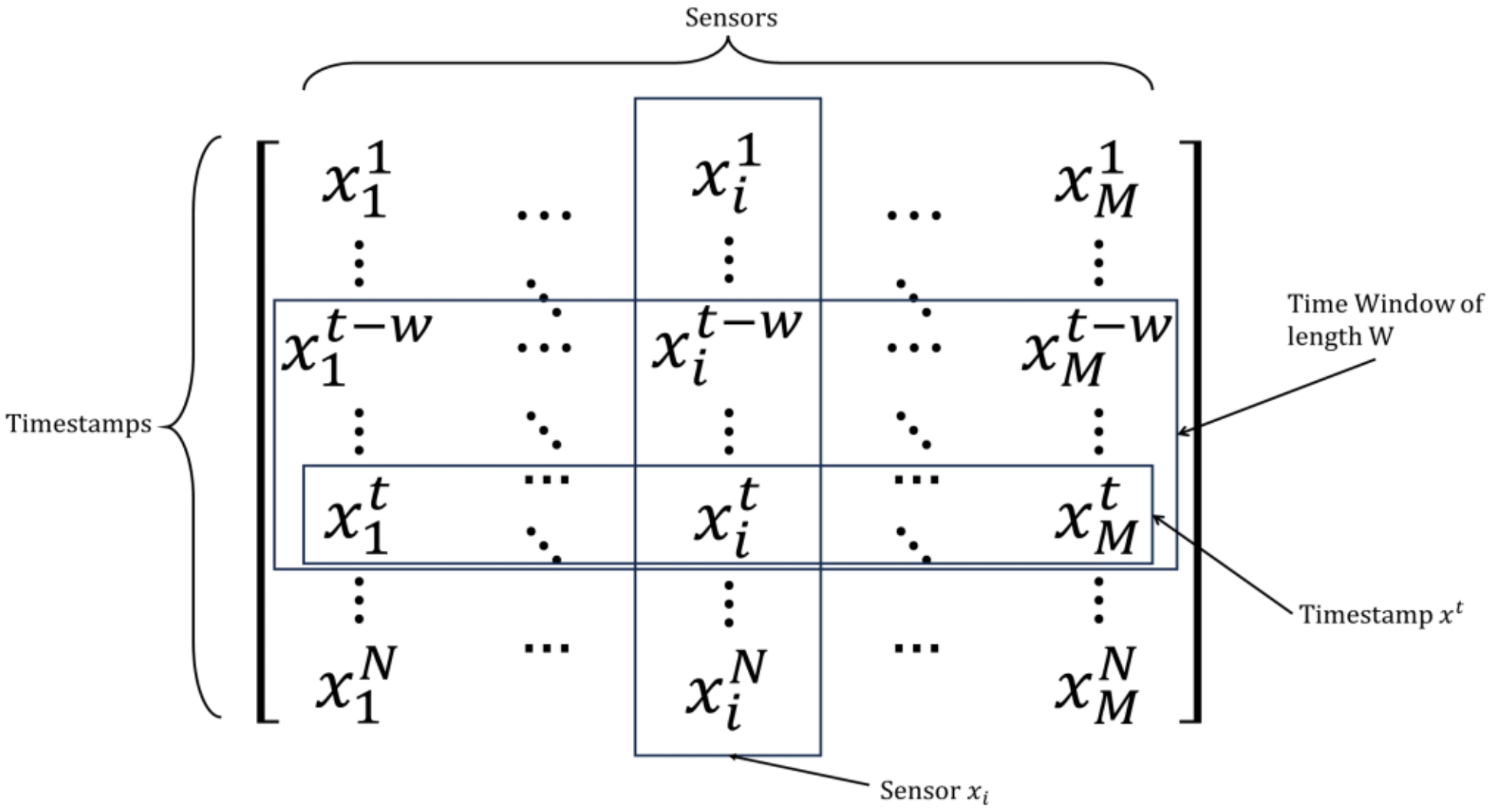

Figure 1: Matrix of multi-sensor time series data, $X \in S^{N\times M}$. Each row $x_i$ is a timestamp sample, and each column $x_i$ represents a sensor reading.

## 2-2- Proposed Method

We propose a hybrid method that combines temporal and spatiotemporal correlations. This way, our method accounts for all types of correlations in MTS data explicitly. Our method also accounts for the stochasticity of time series using variational inference.

Figure 2 shows the framework of the proposed method called HEADMTS. The first component preprocesses the MTS to make time windows. Then, the windows are fed to the temporal correlation component (TC) and spatiotemporal correlation component (STC), to extract all types of correlation explicitly. The TC component uses 2-dimensional Convolutional Neural Networks (CNN) and Gated Recurrent Unit (GRU) to extract temporal information. The STC component uses 2 dimensional CNN and variational inference to model the distribution of MTS and its stochasticity. The HEADMTS uses a hybrid loss function in the end to detect anomalies. After the training stage, we have a well-trained model on the training data, $\phi$ . This model is used to generate a threshold $\delta$. The threshold is used to detect anomalies in MTS in the final stage. We will explain in more detail in the next sections.

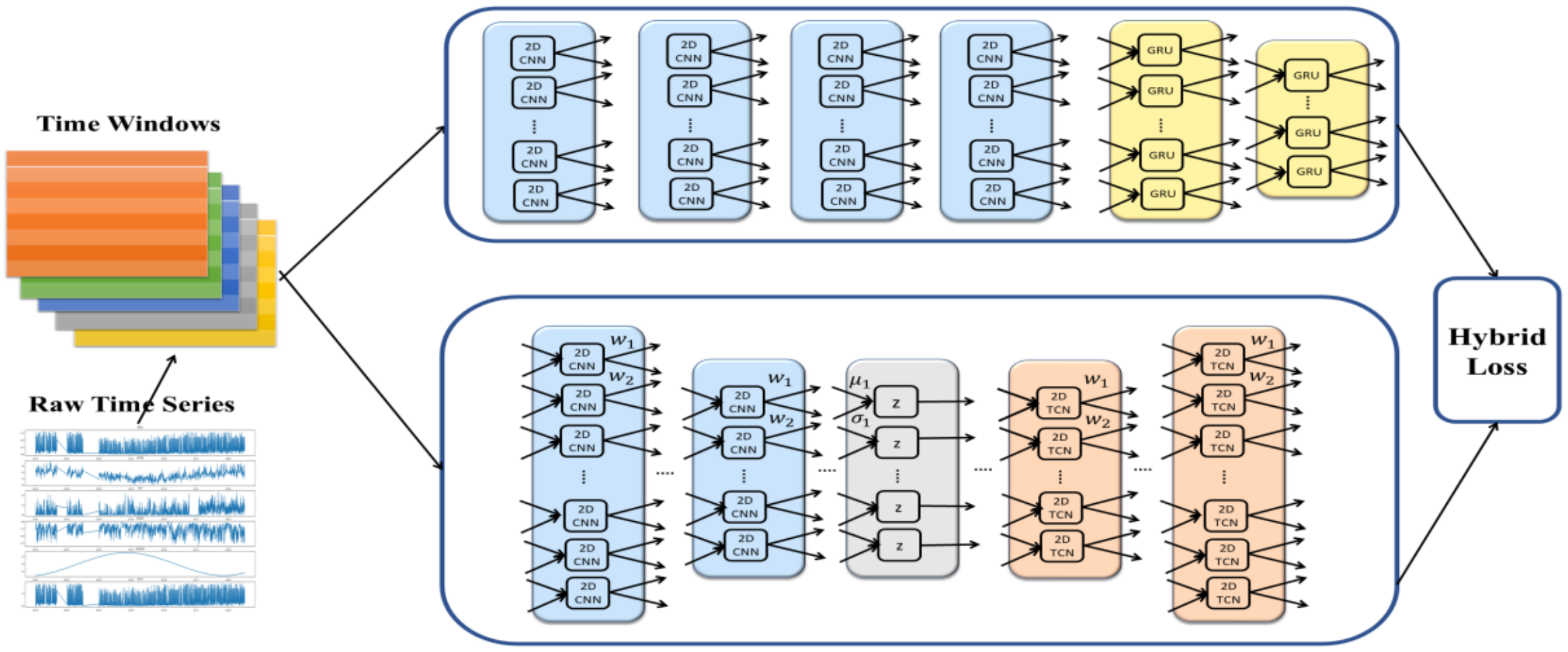

Figure 2: Architecture of HEADMTS

## 2-3- Training Phase

### 2-3-1- TC (Temporal Correlation) Component

In this component, we use two types of neural networks, 2d-CNN and GRU. We deal with multivariate time series. To capture intercorrelation properly, we propose to use 2d-CNN. 2d-CNN building blocks used in our method contain three kinds of layers: 2d convolutional layer, batch normalization layer, and leakyRelu layers. All three layers are wrapped in a time-distributed layer so that the final output can be fed to GRU layers. The 2d convolutional layer maps the input into a feature space containing a correlation of different sensor data, called a feature map, using a sliding filter. In this work, the input of the 2d-CNN block is a time window of the original features. The operation in the convolutional layer $l$ is formulated as follows:

$$y_l^j = \sum_{i \in K_j} w_l^{i,j} \otimes x_{j:|w|}^i + b_{j:|w|}^j \quad (2)$$

$$z_l^{j:|w|} = Relu(y_l^j) \quad (3)$$

where $z_l^{j:|w|}$ is $j$-th feature map of the layer $l$, $K_j$ is the set of 2d filters, $y_l^j$ is the output of the convolution operation, $w_l^{i,j}$ is the 2d convolution kernel $i$ and $b_l^j$ is the bias. Each 2d kernel(filter) creates a correlation map of the sliding kernel. Then, a rectified linear activation function is used.
In the batch normalization layer, we use the following operation to normalize the output of the 2d layer:

$$z_l' = \left(\frac{z - m_z}{s_z}\right).\beta + \rho \quad (4)$$

Where $z_l'$ is the output of the batch-normalization layer, $m_z$ is the mean of the output of the 2d-CNN layer, $s_z$ is the standard deviation and $\beta$ and $\rho$ are learnable parameters of the layer, which, in the end, shift the mean and standard deviation of the output distribution. The output distribution would have a mean of $\rho$ and a standard deviation of $\beta$.
The last layer of the 2d-CNN block is the leakyRelu layer. This layer helps to deal with gradient vanishing/exploding problems while keeping the capability of the deep neural network for modelling non-linearity.
Each 2d-CNN block is wrapped in a TimeDistributed layer so that 2d-CNN operations can be applied to all timestamps.

After deriving convoluted features via several layers of the 2d-CNN blocks, the TC components use GRU for extracting temporal correlation from the extracted feature maps. GRU and Long-Short Term Memory (LSTM) are the most used versions of RNN models due to their power in dealing with gradient vanishing/exploding problems, which was a great gap in vanilla RNNs. We use GRU rather than LSTM since it has a smaller number of gates. This means less complexity and is less prone to overfitting issues. The GRU cells contain two types of gates, the update gate $z$ that decides to update the output of the current state via $\tilde{h}$ or not, and the reset gate $r$ that decides to ignore the previous hidden state or not, thereby controlling the amount of data that should be remembered from previous timestamps. We express the calculations as follows:

$$z_t = \sigma(W_z \cdot x_t + U_z \cdot h_{t-1} + b_z) \tag{5}$$

$$r_t = \sigma(W_r \cdot x_t + U_r \cdot h_{t-1} + b_r) \tag{6}$$

$$\tilde{h}_t = \tanh(W_h \cdot x_t + r_t * U_z \cdot h_{t-1} + b_h) \tag{7}$$

$$h_t = z_t \odot h_{t-1} + (1 - z_t) \odot \tilde{h}_t \tag{8}$$

where $z, r,$ $\tilde{h}$ and $h$ are outputs of the update gate, reset gate, activation function, and output gate, respectively. $\sigma$ is the sigmoid activation function and $tanh$ is the hyperbolic tangent activation function. We use the mean squared loss as the optimization function, between the prediction and the real value as follows:

$$\mathcal{L}_{TC} = \left\| X^t - \hat{X}^t \right\|_2^2 \tag{9}$$

**2-3-2- STC(Spatiotemporal Correlation) Component**

In this component, we develop an auto-encoder to capture the spatiotemporal correlations of the MTS. Using reconstruction information rather than prediction information of the MTS helps us to better focus on spatial correlation. This component learns the latent distribution of MTS through variational inference and then reconstructs the input. Again, our building block is 2d-CNN for encoding the input to a latent distribution, and then we use 2d-CNNTranspose blocks to reconstruct the input from the learned latent distribution. Using a probabilistic distribution helps to capture the stochasticity of the MTS as well.

The vanilla auto-encoders transform the input into a lower-dimension representation (a latent vector) and then reconstruct the input from this latent representation. The main issue of such an approach is that it is prone to learning the identity function rather than the real representation. The other issue is that it encodes each sample of the data regardless of the relation with other samples, therefore ignoring any underlying distribution. Another important aspect is the relation between different time series, since our data is MTS. To deal with these issues, we use 2d convolutional blocks as the building blocks of the encoder and 2d convolutional transpose blocks as the decoder. This makes the model capable of capturing correlations between sensors. To address the latter issue, we use variational inference to learn the probabilistic distribution of the input via the latent distribution and capture the stochasticity of the time series.

In the STC component, the probabilistic encoder is denoted by $p(z|x)$ and the probabilistic decoder is denoted by $p(x|z)$ via which the reconstruction is done. $z$ is the latent variable which represents the latent distribution of the input. We consider a prior distribution for $z$ denoted by $p(z)$. This prior distribution serves as a regularization, which will be detailed soon. The relation between these terms is expressed by Bayes theorem as follows:

$$p(z|x) = \frac{p(x|z)p(z)}{p(x)} = \frac{p(x|z)p(z)}{\int p(x|c)p(z)\, dc} \tag{10}$$

where $p(z|x)$ is called posterior distribution and $p(x|z)$ is the likelihood. The major issue in the calculation of the above formula is the intractable integral in the denominator. That is why we use variational inference to approximate the integral. The intuition is to consider a set of parameterized families of distributions and look for the best approximation among them. We define $p(x|z)$ and $p(z)$ as follows:

$$p(z) = \mathcal{N}(0, I) \tag{11}$$

$$p(x|z) = \mathcal{N}(f(z), I) \tag{12}$$

where $f(z)$ is the mean of the distribution defined by the decoder. Since it is intractable to compute $p(z|x)$, We approximate it by a Gaussian distribution, which is expressed as follows:

$$q_x(z) = \mathcal{N}(\mu_x = g(x), \sigma_x = h(x)) \tag{13}$$

where $g(x)$ and $h(x)$ are mean and variance of $q_x(z)$. We consider G and H as the space for $g(x)$ and $h(x)$ respectively. We use Kullback-Leibler divergence to measure the distance between $q_x(z)$ and $p(z|x)$. Therefore, we aim to optimize the following expression:

$$\begin{aligned}(g^*, h^*) &= \underset{(g,h)\in G\times H}{\operatorname{argmin}} KL(q_x(z), p(z|x)) \\ &= \underset{(g,h)\in G\times H}{\operatorname{argmin}} \left( \underset{z\sim q_x}{\mathrm{E}} (\log q_x(z)) - \underset{z\sim q_x}{\mathrm{E}} \left( \log \frac{p(x|z)p(z)}{p(x)} \right) \right) \\ &= \underset{(g,h)\in G\times H}{\operatorname{argmax}} \left( \underset{z\sim q_x}{\mathrm{E}} (\log p(x|z)) - KL(q_x(z), p(z)) \right) \\ &= \underset{(g,h)\in G\times H}{\operatorname{argmax}} \left( \underset{z\sim q_x}{\mathrm{E}} \left( -\frac{\|x - f(z)\|^2}{2c} \right) - KL(q_x(z), p(z)) \right)\end{aligned} \tag{14}$$

Where $f(z)$ is needed to be optimized so that it maximizes the likelihood $p(x|z)$. Therefore, in the end, we need to optimize the following expression:

$$(f^*, g^*, h^*) = \underset{(f,g,h)\in F\times G\times H}{\operatorname{argmax}} \left( \underset{z\sim q_x}{\mathrm{E}} \left( -\frac{\|x - f(z)\|^2}{2c} \right) - KL(q_x(z), p(z)) \right) \tag{15}$$

We model $f^*$ via decoder and $g^*$ and $h^*$ via encoder networks using 2d-CNN and 2d-CNNTranspose blocks. For both the encoder and decoder, we use a Gaussian distribution. $g^*$ and $h^*$ are the mean and covariance of the encoder and $f^*$ is the mean of the decoder with constant covariance. After the encoding step and approximating $q_x(z)$, We need to sample from this distribution to derive $f(z)$ and reconstruct the input. But z is a random variable, which makes the differentiation and gradient descent impossible, therefore making the backpropagation impossible. To address this issue, we use the reparameterization trick [12] as follows:

$$z \sim q_x(z) = \mathcal{N}(\mu_x, \sigma_x) \rightarrow z = \mu_x + \gamma * \sigma_x \ ; \ \gamma \sim \mathcal{N}(0, I) \tag{16}$$

Finally, the loss function used for the STC component is expressed as follows:

$$\mathcal{L}_{STC} = \|x - f(z)\|^2 + KL(q_x(z), p(z) = \mathcal{N}(0, I)) \tag{17}$$

### 2-3-3- Detection Component

This component combines the $\mathcal{L}_{TC}$ and $\|X - f(z)\|^2$ errors over validation data $\langle \mathrm{X}_v, \Upsilon_v \rangle$ to create the threshold $\delta$. We denote the threshold as follows:

$$\delta^* = \underset{(X, X^t)\in \mathrm{X}_v\times\Upsilon_v}{\operatorname{argmin}} \left( \lambda \|X - f(z)\|^2 + (1 - \lambda) \left\| X^t - \widehat{X}^t \right\|_2^2 \right) \tag{18}$$

where $\lambda$ is a trade-off hyperparameter between $\|X - f(z)\|^2$ and $\|X^t - \hat{X}^t\|_2^2$.

HEADMAS uses $\delta^*$ to do the detection. We have test data denoted as $D_{ts} = \langle X_{ts}, Y_{ts} \rangle$ . HEADMTS model calculates the hybrid error for each test data sample as follows:

$$E_{ts} = \left\{ e_i = \lambda \|X - f(z)\|^2 + (1-\lambda)\|X^t - \hat{X}^t\|_2^2 \middle| \langle X_i^{ts}, {X^t}_i^{ts} \rangle \in D_{ts} \right\} \quad (19)$$

Then, the detection function, denoted by $\phi(X_i^{ts})$ is carried on as follows:

$$\phi(X_i^{ts}) = \begin{cases} 0 & if\ e_i \leq \delta \\ 1 & o.w \end{cases} \quad (20)$$

The HEADMTS is shown in Algorithm 1.

# 3- Experiments

## 3-1- Experimental Setup

We use Python 3.7.6 in the Spyder 4.0.1 platform. We use Tensorflow 1.14.0 and the Keras library to develop the neural network. The configuration of baseline methods is explained in Table 1.

| **Algorithm 1** | Training and Detection Procedure of HEADMTS |
|---|---|
| | **Training Phase** |
| **Input** | MTS data $X$, batch size $M$, window size $W$ |
| **Output** | model parameters $W_{TC}$ and $W_{STC}$ |
| **1:** | Convert raw MTS into windows $X_{tr} = \{x_{1:M}^{1:w}, \dots, x_{1:M}^{t-w:t}, \dots, x_{1:M}^{N-w:N}\}$ , and targets $Y_{tr} = \{x_{1:M}^{w}, \dots, x_{1:M}^{t}, \dots, x_{1:M}^{N}\}$ |
| **2:** | **Do:** |
| **3:** | Calculate feature maps for $x_{1:M}^{t-w:t}$ via 2d-convolutional network;//Eq (2) and (3) |
| **4:** | Do batch-normalization;//Eq (4) |
| **5:** | Calculate $Y_i$ through GRU network with 64 and 32 neuron layers;//Eq (5) to (8) |
| **6:** | Calculate the $\mathcal{L}_{TC}$, $\frac{\partial \mathcal{L}_{TC}}{\partial w}$ and update $W_{TC}$;//Eq (9) |
| **7:** | **While** not converged |
| **8:** | **Do:** |
| **9:** | Calculate feature maps for $x_{1:M}^{t-w:t}$ via 4 2d-convolutional network networks with 32, 64, 64, and 64 neurons;//Eq (2), Eq (3) |
| **10:** | Do batch-normalization;//Eq (4) |
| **11:** | Calculate to approximate $(g^*, h^*) = \underset{(g,h)\in G\times H}{\operatorname{argmax}} \left( \underset{z\sim q_x}{\mathrm{E}} \left( -\frac{\|x-f(z)\|^2}{2c} \right) - KL\big(q_x(z), p(z)\big) \right)$; |
| **12:** | Sample $z$ via reparameterization trick, $z = \mu_x + \gamma * \sigma_x$ ; $\gamma \sim \mathcal{N}(0, I)$; |
| **13:** | Calculate $f^*$ via 4 2d-CNNTranspose;// |
| **14:** | Calculate the $\mathcal{L}_{STC}$, $\frac{\partial \mathcal{L}_{STC}}{\partial w}$ and update $W_{STC}$ |
| **15:** | **While** not converged |
| **16:** | **Return** $W_{pre}$ and $W_{rec}$ |
| | **Detection Phase** |
| **Input:** | test data |
| **Output:** | Threshold $\delta$, Labels $\varphi(X_i)$ for all $X_i \in \{X_1, \dots, X_N\}$ |
| **1:** | Calculate $\delta^* = \underset{(X,X^t)\in X_v\times Y_v}{\operatorname{argmin}} \left( \lambda\|X - f(z)\|^2 + (1-\lambda)\|X^t - \hat{X}^t\|_2^2 \right)$;// Eq |
| **2:** | Transform MTS time series into windows $X_{ts} = \{x_{1:M}^{1:w}, \dots, x_{1:M}^{t-w:t}, \dots, x_{1:M}^{N-w:N}\}$, and targets $Y_{ts} = \{x_{1:M}^{w}, \dots, x_{1:M}^{t}, \dots, x_{1:M}^{N}\}$ |
| **3:** | **For** $X_i \in X_{ts}; i = 1, \dots, N$ **do**: |
| **4:** | Calculate $\hat{Y}_i$ via the TC network; |
| **5:** | Calculate $\hat{X}_i$ via the STC network; |

Calculate hybrid loss $\mathcal{L}_{X_i} = \lambda \|X - f(z)\|^2 + (1-\lambda)\|X^t - \hat{X}^t\|_2^2$;
**IF** $\mathcal{L}_{X_i} > \delta^*$:
$\phi(X_i) = "Anomaly"$
**Else:**
$\phi(X_i) = "Normal"$
**End for**
**Return** Labels $\phi(X_i)$ for all $X_i \in \{X_1, \dots, X_N\}$

Table 1: Configuration of baselines

| | Baseline Methods for Comparison |
|---|---|
| LOF | Number of neighbors:10 |
| IF | Number of estimators:100, contamination:0.01 |
| OCSVM | Kernel:"rbf", gamma:0.6 |
| LSTM-NDT | Number of LTSM layer: 2, Neurons:80, Dropout:0.3, loss:mse, solver=Adam, learning rate (lr):1e-3 |
| MSCRED | Encoder layer sizes={32,64,128,256}, Decoder layer sizes={256,256,128,64}, ConvLSTM Cell:dimension=2,kernel shape=[2,2],forget-bias=1.0; Learning rate:0.0002;threshold:0.005 |
| MAD-GAN | Latent Dimension:15;Number of Variables:78;Learning Rate:0.1; L2 Norm Bound:$1e^{-5}$;dp sigma:$1e^{-5}$; dp:False |
| OmniAnomaly | RNNCell:GRU; Hidden Dimension:500;Window length:20;l2 Regulizer:0.0001;Search Min:-1;Search Max:1;Level:0.01;NF Layer:20 |
| USAD | Hidden Size:100; Loss:Relu; Window length:20; |
| VAEpro | Weight Decay:$1e^{-3}$;Dropout:0.2;Activation:Relu;Number of Layers:5 |

We run experiments using baselines, including LOF [13], IF [14], OCSVM [15], LSTM-NDT [2], MSCRED [10], MAD-GAN [1], OmniAnomaly [11], USAD [3], VAEpro [7].
For the MTS dataset, we use three widely used datasets SWAT [1], WADI [1], and SMD [11].

## 3-2- Evaluation Criteria

To evaluate the performance of methods, we use four widely used metrics. The metrics are defined as follows:

$$Precision = \frac{TP}{TP + FP} \tag{21}$$

$$Recall = \frac{TP}{TP + FN} \tag{22}$$

$$F1 = 2 \times \frac{Precision \times Recall}{Precision + Recall} \tag{23}$$

where TP is the true positive, FP is the false positive, and FN is the false negative. Of these three metrics, F1 is the most important one. It finds a balance between precision and recall. As the Precision formula suggests, it penalizes false positives, while the recall penalizes false negatives. In smart meter systems, we believe that a false negative is more costly than a false positive. Therefore, after the f1 measure, recall is more important than precision since it penalizes false negatives.

## 3-3- Results & Discussion

In this section. We discuss the results of the performance comparison in terms of the mentioned metrics. We also show the convergence rate and run an ablation study as well, to show the effectiveness of each component. Table 2 shows the final configuration of the neural network used for the training stage model.

Table 2: Configuration of neural network models used in HEADMTS.

| | Layers | Neurons | Filters | Kernel | Stride | Regularizer | Activation Function |
|---|---|---|---|---|---|---|---|
| TC Component | 4 2D-Convolution Blocks | 32<br>64<br>64<br>64 | 32<br>64<br>64<br>64 | (3,3) | 1 | Kernel Regularizer =L1L2 BiasRegulaizer=L2 | LeakyRelu |
| | GRU 1<br>GRU 2 | 64<br>32 | | | | Dropout=2.0 | Linear |
| STC Component | 4 2D-Convolution Blocks | 32<br>64<br>64<br>64 | 32<br>64<br>64<br>64 | (3,3) | 1 | Kernel Regularizer =L1L2 BiasRegulaizer=L2 | LeakyRelu |
| | 4 2D-Convolution Transpose Blocks | 64<br>64<br>64<br>32 | 64<br>64<br>64<br>32 | (3,3) | 1 | Kernel Regularizer =L1L2 BiasRegulaizer=L2 | LeakyRelu |

Table 3:Performance in terms of Accuracy, Precision, Recall, and F1

| Baselines | SWaT | | | WADI | | | SMD | | | |
|---|---|---|---|---|---|---|---|---|---|---|
| | Precision | Recall | F1 | Precision | Recall | F1 | Precision | Recall | F1 | Avg F1 |
| LOF | 0.3208 | 0.5045 | 0.3926 | 0.4839 | 0.5046 | 0.4921 | 0.4211 | 0.2824 | 0.3422 | 0.4089 |
| IF | 0.4108 | 0.4737 | 0.4111 | 0.4311 | 0.1903 | 0.2108 | 0.4111 | 0.2512 | 0.2624 | 0.2947 |
| OCSVM | 0.3112 | 0.4832 | 0.3802 | 0.4732 | 0.4413 | 0.4645 | 0.4461 | 0.4011 | 0.7322 | 0.5256 |
| LSTM-NDT | 0.5901 | 0.7070 | 0.6433 | 0.4411 | 0.5952 | 0.5067 | 0.7111 | 0.8364 | 0.7687 | 0.6395 |
| MSCRED | 0.4302 | 0.5749 | 0.4921 | 0.5311 | 0.5636 | 0.5469 | 0.7911 | 0.8623 | 0.8252 | 0.6214 |
| MAD-GAN | 0.5601 | 0.5274 | 0.5433 | 0.4811 | 0.3119 | 0.3785 | 0.9014 | 0.8918 | 0.8966 | 0.6061 |
| OmniAnomaly | 0.3605 | 0.9892 | 0.5300 | 0.3111 | 0.5323 | 0.3927 | 0.8891 | **0.9891** | **0.9428** | 0.6218 |
| USAD | **0.6605** | 0.7778 | 0.7144 | 0.4711 | 0.3912 | 0.4275 | 0.8811 | 0.9247 | 0.9024 | 0.6814 |
| VAEpro | 0.6311 | 0.9244 | 0.7501 | 0.7431 | 0.8432 | 0.7900 | 0.7851 | 0.8371 | 0.8103 | 0.7834 |
| HEADMTS | 0.6401 | **0.9994** | **0.7804** | **0.7945** | **0.9921** | **0.8824** | **0.9145** | 0.9483 | 0.9311 | **0.8646** |

Table 3 shows the advantages of our method against baselines. The most important metric we rely on is the Average F1 measure shown in the last column. As the table shows, our method, HEADMTS, outperforms all other baselines by a considerable margin, with F1 of 0.8646. It is due to the ability of our method to consider both temporal and spatiotemporal correlation between features, and using 2d-CNN to capture all types of relations explicitly gives an advantage to HEADMTS. The second-best method is VAEpro. It uses variational inference as well to capture stochasticity, but it only accounts for spatial relations between features. It also uses MLP as the building block of its neural network, which is not as effective as 2d-CNN in both capturing non-linearity and correlation between features. IF, LOF, and OCSVM show the poorest performance. These methods don't capture temporal information. They also suffer from the curse of dimensionality; therefore, they can't model the relationship between features properly.

The most difficult dataset among the three mentioned MTS datasets is WADI. It has 124 sensor time series that have complicated relationships. The result in the table shows that HEADMTS outperforms all other baselines to a good extent. It shows the advantage of HEADMTS in capturing correlations. The F1 of our method for WADI is 0.8824. It is again followed by VAEpro. The worst performance is shown by IF. The existence of complex relations between sensor time series makes it difficult for IF to show good performance. MAD-GAN also shows poor performance. It is an adversarial-based model that uses LSTM as the building block. LSTM is good at capturing temporal information, but might not be as good as 2d-CNN for capturing spatiotemporal correlations. The other reason for its poor performance can be using dimensionality reduction techniques before feeding the MTS to the model, which causes the loss of valuable information associated with the relation between features. The other reason can be mode collapse, which is an issue associated with adversarial-based models.

In terms of recall for WADI, HEADMTS outperforms with a recall of 0.9921. It shows that HEADMTS has low false negatives, which is a very good point for monitoring sensitive systems like energy systems. The second-best recall for WADI is shown by VAEPro. The worst performances belong to IF and MAD-GAN. In terms of precision, HEADMTS outputs a precision of 0.7945 followed by VAEPro with 0.7431. The worst performance is shown by Omnianomaly with 0.3111. It is a reconstruction-based method that uses GRU as the building blocks. It only considers spatial relationships for which it uses GRU, a neural network designed for capturing temporal information.

The second-most difficult dataset is SWAT with 78 sensor time series. Again, here we observe that HEADMTS outperforms other baselines with 0.7804. It shows that HEADMTS has superiority when there are complicated relationships between sensor time series. The second-best is again VAEPro with the competitive result of 0.7501. VAEPro is like our method. The worst results are shown by OCSVM and LOF, with 0.3802 and 0.3926, respectively.

In terms of recall, HEADMTS is superior with 0.9994, which is closely followed by Omnianomay with 0.9892. While Omnianomaly wasn't good in very high-dimensional datasets like WADI, as the number of sensors starts to decrease, Omnianomaly shows its power using adversarial generative ideas. In terms of precision, USAD is superior with a precision of 0.6605, which is closely followed by HEADMTS with a precision of 0.6401. USAD shows a relatively good F1 measure of 0.7144. It is an auto-encoder-based model that uses reconstruction error to model the training data. It combines the ideas of auto-encoders with adversarial learning. They didn't use variational inference. Instead, the encoding and decoding process of USAD is more deterministic than the stochastic approach taken by HEADMTS. USAD also doesn't focus on capturing temporal correlations as well. The worst precision is shown by OCSVM and LOF, with 0.3112 and 0.3201, respectively.

The last dataset is SMD, which has 39 sensor time series. It is less complex than the other two datasets, due to a simpler relation between features. In this dataset, Omnianomaly is superior by a very slight margin to HEADMTS, with 0.9428 over 0.9311. As mentioned above, Omnianomaly gains an edge when the correlation between features becomes less complex. It shows adversarial learning can be more effective when the relation between features is simpler. Otherwise, taking all kinds of correlation is crucial as HEADMTS does. LOF and IF again are among the worst performance results, with 0.3422 and 0.2624. However, other baselines show better performance than the other two datasets. It shows the superiority of considering non-linearity by deep learning methods over LOF and IF. Also, the ability of deep learning to deal with the curse of dimensionality is better shown in this dataset.

In terms of recall, Omnianomaly is ahead of our method by a better margin, with 0.9891 over 0.9493. Still, HEADMTS is the second-best and is competitive with Omnianomaly. LOF and IF show the worst recall with 0.2824 and 0.2512. In terms of precision, HEADMTS is superior, with 0.9145, followed by MAD-GAN with 0.9014. MAD-GAN is another adversarial-based model. It shows that adversarial-based models are successful in detecting anomalies for SMD datasets. But still, the performance of HEADMTS is competitive in SMD. The worst performance in terms of precision is shown by LOF and IF, with 0.4211 and 0.4111, respectively.

In the end, the consideration of temporal and spatiotemporal correlations with TC and STC components, through 2d-CNN explicitly combined with GRU and variational inference ideas, shows considerable advantage for detecting anomalies in complex time series datasets.

Figure 3 shows the fast convergence of HEADMTS components after almost 10 epochs. The left curve belongs to the TC component, and the right curve belongs to the STC component. A proper number of layers, the right choice of activation functions like leakyRelu, and the layer-level regularisations are the reasons behind this fast convergence. Choosing 2d-CNN for MTS data is another reason why the components converge that fast.

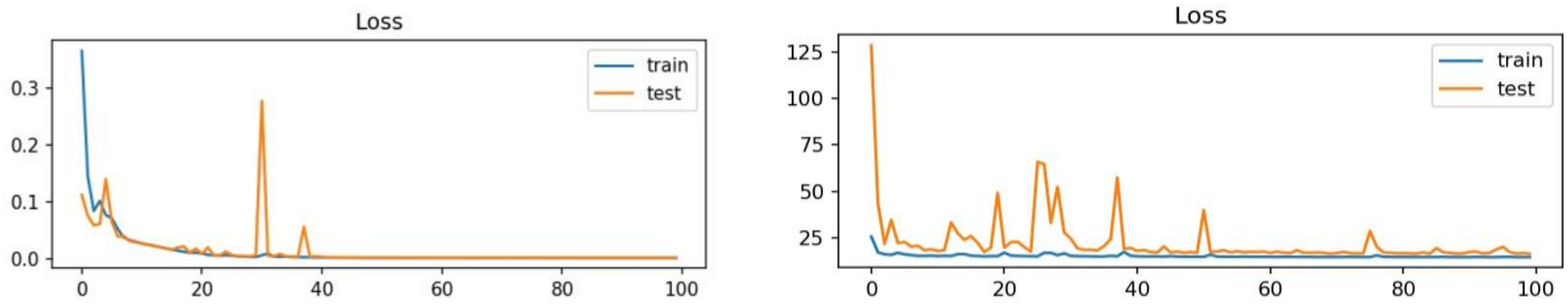

Figure 3: Convergence rate of TC and STC components

In the next study, we run an ablation study to assess the effectiveness of each component. We create two variants of HEADMTS, namely tcHEADMTS and stcHEADMTS. The former only uses the TC component and its associated loss to detect anomalies, and the latter uses only the STC component. We run the study for all three datasets. The results are shown in Table 4.

Again, in terms of the most important metric, which is the average F1, we observe that HEADMTS is superior with an average F1 of 0.8646. Still, we observe that both components show considerable performance, with 0.8422 for tcHEADMTS and 0.8320 for stcHEADMTS. These results show that tcHEADMTS and the associated TC component are more effective than the STC component. This is mainly due to the existence of 2d-CNN in the TC component. Using 2d-CNN in combination with GRU helps to capture both types of correlations, temporal and spatial. While STC only focuses on spatial correlations. Still, their combination yields better results.

For the WADI dataset, which is the most difficult dataset, again HEADMTS is superior, followed by stcHEADMTS in the second place with 0.8621. This is mainly due to the presence of complicated spatial correlations in this dataset and the better capability of the STC component in capturing spatial correlations. In terms of recall, again HEADMTS has the best performance with 0.9921. The second place again belongs to stcHEADMTS with 0.9612. In terms of precision, again HEADMTS outperforms with a precision of 0.7945 followed by tcHEADMTS with 0.7925.

The second dataset in terms of complexity of spatial correlations is SWAT. In this dataset, HEADMTS outperforms in terms of F1 with 0.7804. This time, second place belongs to tcHEADMTS. This is due to less spatial complexity and higher complexity in terms of temporal correlations. tcHEADMTS is better at capturing temporal correlations. In terms of recall, HEADMTS is superior with 0.7338, and stcHEADMTS is ahead of tcHEADMTS with a slight margin. In terms of precision, tcHEADMTS shows the best performance with 0.8449, with HEADMTS owning the second place with 0.8332. It shows that tcHEADMTS has fewer false positives but more false negatives.

The last dataset, SMD has more complicated temporal correlations than spatial correlations. Again, in terms of F1, HEADMTS is the best one with 0.9311, and tcHEADMTS is the second-best with 0.9102. It shows that in the presence of temporal correlations, the TC component is more effective than the STC component. In terms of recall, HEADMTS is superior with 0.9483, and tcHEADMTS is second with 0.9194. The same ranking is observed for precision, with HEADMTS the best and tcHEADMTS the second. We observed that the combination of two components yields better results, and the TC component is more effective than the STC component.

Table 4: ablation study

| Variants | SWaT | | | WADI | | | SMD | | | Avg F1 |
|---|---|---|---|---|---|---|---|---|---|---|
| | Precision | Recall | F1 | Precision | Recall | F1 | Precision | Recall | F1 | |
| tcHEADMTS | **0.8449** | 0.7108 | 0.7721 | 0.7925 | 0.9033 | 0.8443 | 0.9011 | 0.9194 | 0.9102 | 0.8422 |
| stcHEADMTS | 0.8311 | 0.7157 | 0.7691 | 0.7815 | 0.9612 | 0.8621 | 0.8814 | 0.8488 | 0.8648 | 0.8320 |
| HEADMTS | 0.8332 | **0.7338** | **0.7804** | **0.7945** | **0.9921** | **0.8824** | **0.9145** | **0.9483** | **0.9311** | **0.8646** |

## 4- Conclusion

In this paper, we studied the problem of anomaly detection in multivariate time series (MTS). This is a demanded and popular problem in modern systems that are monitored by smart meters. Multiple sensors together collect information on system status and help maintain the health of such systems. But this problem raises several challenges, like the complexity of relation between different time series associated with different sensors. Also, the lack of labelled data makes this problem an unsupervised learning problem, which is another challenge. To detect anomalies in such systems, a method needs to capture all types of correlation, including temporal and spatiotemporal information. We proposed a hybrid method that captures both these correlations explicitly. Our method consisted of two components, TC and STC. The former uses 2D-CNN and GRU to model the temporal information, and the latter uses a variational autoencoder, whose building block is 2D-CNN, which is powerful at capturing spatio-temporal correlations. We conducted several experiments against state-of-the-art, and our method, called HEADMTS, outperformed state-of-the-art in terms of F1 measure. We showed the fast convergence rate of HEADMTS. We also conducted an ablation study to show the effectiveness of each component. The empirical experiments endorse our ideas that capturing both correlations explicitly and 2D-CNN are effective at detecting anomalies in MTS.